\documentclass{article} 
\usepackage{iclr2024_conference,times}

\usepackage{amsmath,amsfonts,bm}

\def\eqref#1{equation~\ref{#1}}

\def\1{\bm{1}}

\DeclareMathAlphabet{\mathsfit}{\encodingdefault}{\sfdefault}{m}{sl}
\SetMathAlphabet{\mathsfit}{bold}{\encodingdefault}{\sfdefault}{bx}{n}

\usepackage{makecell}
\usepackage[pagebackref=true,breaklinks=true,citecolor=blue,linkcolor=blue,colorlinks,urlcolor=blue,bookmarks=false]{hyperref}
\usepackage{url}
\usepackage{booktabs}
\usepackage{arydshln}
\usepackage{multirow}
\usepackage{graphicx}
\usepackage{subcaption}
\usepackage{enumitem}
\usepackage{xspace}
\usepackage{mathabx}
\usepackage{bbm}
\usepackage{wrapfig}
\usepackage{xcolor}
\usepackage{bbm}

\graphicspath{{figures/}}

\usepackage[capitalize]{cleveref}
\crefname{section}{Section}{Secs.}
\Crefname{section}{Section}{Sections}
\Crefname{table}{Table}{Tables}
\crefname{table}{Tab.}{Tabs.}

\iclrfinalcopy

\title{Open-MAGVIT2: \\An Open-Source Project Toward Democratizing Auto-regressive Visual Generation}

\author{
\hspace{-3mm}
\noindent
\textbf{Zhuoyan Luo}$^{1,2}$\thanks{Equal Contribution. Work done during an internship at ARC Lab, Tencent PCG.}~\quad \hfill
\textbf{Fengyuan Shi}$^{1,3*}$\quad \hfill
\textbf{Yixiao Ge}$^{1}$\thanks{Corresponding author and project lead.}~\quad \hfill
\textbf{Yujiu Yang}$^{2}$\quad
\textbf{Limin Wang}$^{3}$\quad \hfill
\textbf{Ying Shan}$^{1}$
\vspace{0.3em}
\\
$^1$ARC Lab, Tencent PCG \hfill
$^2$Tsinghua University \hfill
$^3$Nanjing University \hfill
\vspace{0.8em}
\\
\quad \quad \quad \quad  \quad \quad \url{https://github.com/TencentARC/SEED-Voken}
\vspace{-5mm}
}

\begin{document}

\maketitle


\begin{figure}[ht]
\vspace{-5pt}
\begin{center}
	\includegraphics[width=\linewidth]{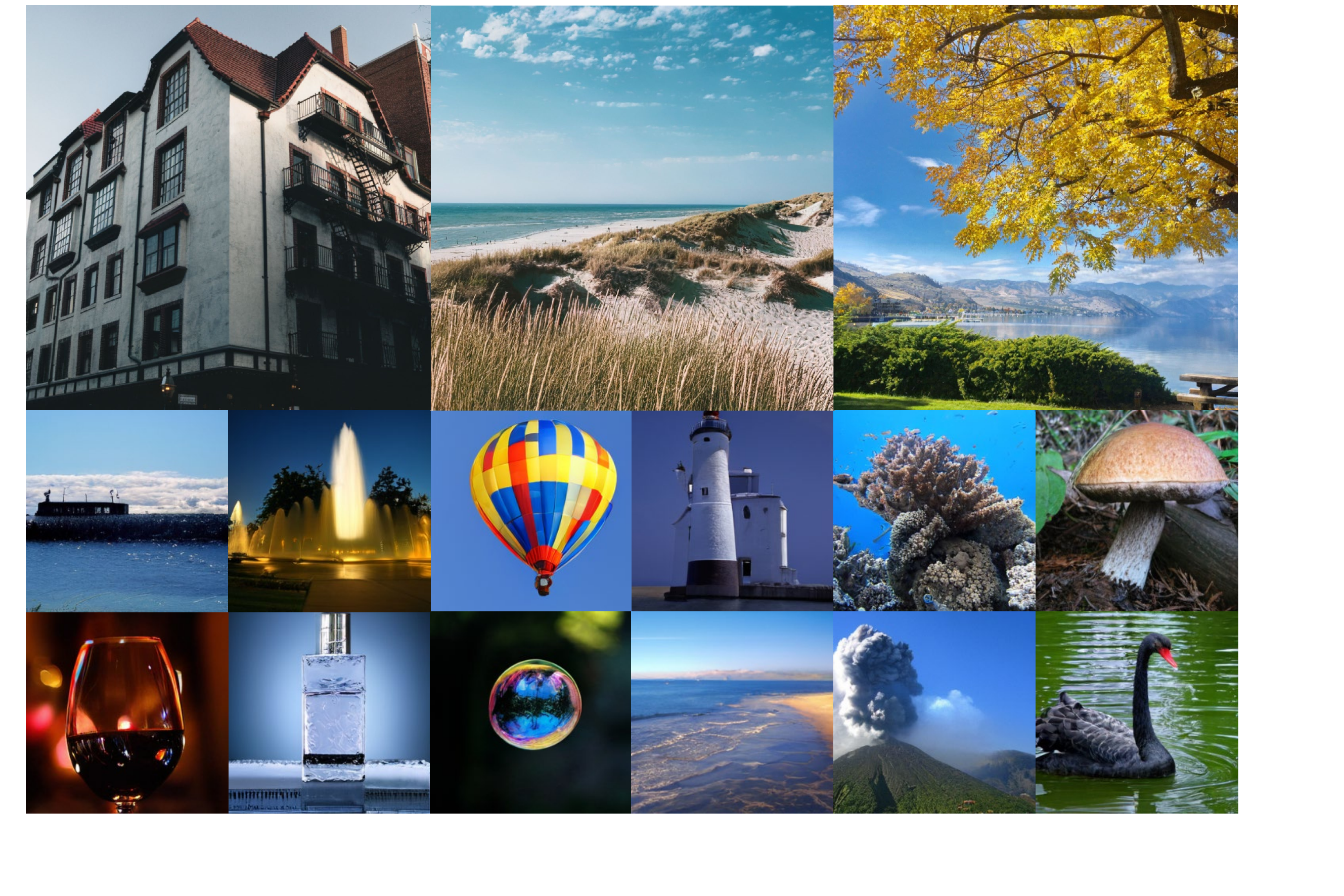}
\end{center}
\vspace{-7pt}
\caption{\small
\textbf{Reconstruction and generation samples of Open-MAGVIT2}. We show $1024\times 1024$ reconstructed samples (top) and $256 \times 256$ generated samples (middle and bottom).
}
\vspace{-5pt}
\label{fig:abs}
\end{figure}


\begin{abstract}


The Open-MAGVIT2 project produces an open-source replication of Google's MAGVIT-v2 tokenizer, a tokenizer with a super-large codebook (i.e., $2^{18}$ codes), and achieves the state-of-the-art reconstruction performance on ImageNet and UCF benchmarks. 
We also provide a tokenizer pre-trained on large-scale data, significantly outperforming Cosmos on zero-shot benchmarks (1.93 vs. 0.78 rFID on ImageNet original resolution).
Furthermore, we explore its application in plain auto-regressive models to validate scalability properties, producing a family of auto-regressive image generation models ranging from 300M to 1.5B.
To assist auto-regressive models in predicting with a super-large vocabulary, we factorize it into two sub-vocabulary of different sizes by asymmetric token factorization, and further introduce ``next sub-token prediction'' to enhance sub-token interaction for better generation quality.  We release all models and codes to foster innovation and creativity in the field of auto-regressive visual generation.

\end{abstract}
\begin{table}[t]
    \centering
    \setlength{\tabcolsep}{4pt}
    \renewcommand\arraystretch{1.0}
    \caption{\textbf{Configurations of visual tokenizers in Open-MAGVIT2 series}. We provide both image and video tokenizers, where the former has a pretrained version to facilitate text-conditional image generation.}
    \resizebox{\linewidth}{!}{
    \begin{tabular}{lccccc}
    \toprule
    \multirow{2}{*}{\textbf{Tokenizer}}  & \textbf{Training} & \multirow{2}{*}{\textbf{Codebook Size}} & \multirow{2}{*}{\textbf{Resolution}} & \textbf{Temporal} & \textbf{Spatial} \\
    & \textbf{Data} & & & \textbf{Ratio} $p_t$ & \textbf{Ratio} $p_s$ \\
    \midrule    
    \multirow{2}{*}{Open-MAGVIT2-I-In1k} & \multirow{2}{*}{ImageNet1k} & \multirow{2}{*}{262144} & $256 \times 256$ & $-$ & $16 \times 16$ \\
    & & & $128 \times 128$ & $-$ & $8 \times 8$ \\ \midrule
    \multirow{2}{*}{Open-MAGVIT2-I-PT} & \multirow{2}{*}{Image-text data} & 16384 & $256 \times 256$ & $-$ & $16 \times 16$ \\
    & & 262144 & $256 \times 256$ & $-$ & $16 \times 16$ \\ \midrule
    Open-MAGVIT2-V-UCF & UCF-101 & 262144 & $128 \times 128$ & 4 & $8 \times 8$ \\
    \bottomrule
    \end{tabular}}
    \label{tab:tokenizer}
\end{table}
\begin{table}[h]
    \centering
    \setlength{\tabcolsep}{4pt}
    \renewcommand\arraystretch{1.0}
    \caption{\textbf{Configurations of auto-regressive image generation models in Open-MAGVIT2 series}. We partially follow the scaling rule proposed in the previous works~\citep{llamagen, var}.}
    \resizebox{\linewidth}{!}{
    \begin{tabular}{lccccc}
    \toprule
    \textbf{Model}  & \textbf{Parameters} & \textbf{Inter-Blocks} $N$ & \textbf{Intra-Blocks} $L$ &  
    \textbf{Widths} $w$  & \textbf{Heads} $h$ \\
    \midrule
    Open-MAGVIT2-AR-B & 343M & 24 & 2 & 1024 & 16 \\
    Open-MAGVIT2-AR-L & 804M & 36 & 3 & 1280 & 20 \\
    Open-MAGVIT2-AR-XL & 1.5B & 48 & 4 & 1536 & 24 \\
    \bottomrule
    \end{tabular}}
    \label{tab:scaling}
\end{table}

\section{Introduction} 

Large Language Models (LLMs), built upon auto-regressive transformer~\citep{attention, openai2023gpt4, palm, llama2}, have demonstrated dominance in natural language generation due to the incredible context modeling and scalability. Inspired by this, emergent works introduce auto-regressive models into visual generation~\citep{vqvae, vqgan, vit-vqgan, rqvae, llamagen}. These approaches first utilize a vector quantizer for visual tokenization and de-tokenization, then employ an auto-regressive transformer for discrete visual token sequence modeling.

Although great processes are achieved, the quality of visual generation still falls behind the diffusion-based methods. The main factor is limited tokenizer performance. Tokenizers are generally posited as the upper bound of the visual generation, and inferior off-the-shelf tokenizers (e.g., VQ-VAE~\citep{vqvae}) will lead to poor generation quality. Although some improvements are done~\citep{vit-vqgan,rqvae,llamagen}, current tokenizers are limited by the codebook size and utilization, and the reconstruction performance is still far worse than VAE\citep{vae, stablediffusion} used in diffusion models. To unlock the potential of tokenizers, MAGVIT-v2~\citep{magvit2} proposes Lookup-Free Quantizer to enable a highly code-activated and super-large codebook, and achieves better generation quality than diffusion models. \textbf{However, such a powerful visual tokenizer is completely closed-source and we have no access to this so far, limiting the development of the academic community.}

In this work, we push forward the auto-regressive visual generation in two folds: 
\begin{itemize}[leftmargin=0.5cm]
    \item \textbf{{Replication of the visual tokenizer}}: We re-implement the advanced Lookup-Free Quantizer proposed by MAGVIT-v2. To our best knowledge, our open-source replication achieves the closest reconstruction performance stated in MAGVIT-v2 (1.18 vs. 1.15 rFID on ImageNet 128$\times$128) and outperforms all other methods on the hallmark Imagenet and UCF benchmarks. Furthermore, we provide the visual tokenizer for text-conditional image generation with improved representational capacity by pretraining it on large-scale general-domain datasets. Our pretrained visual tokenizer achieves state-of-the-art zero-shot reconstruction performance on COCO and ImageNet, surpassing Cosmos~\citep{cosmos} and Llamagen~\citep{llamagen} significantly (Table~\ref{tab:pretrain_recon}).

    \item \textbf{{Integrating a super-large codebook with AR visual generation}}: Instead of simply following MAGVIT-v2 that leverages the vision-oriented design (i.e., mask generative methods~\citep{maskgit} for visual synthesis), we seek to exploit the potential of such a large codebook in vanilla auto-regressive generation. To assist auto-regressive models in predicting with a super-large vocabulary, we factorize it into two sub-vocabulary of different sizes by asymmetric token factorization, and further introduce ``next sub-token prediction'' to enhance sub-token interaction for better generation quality. Our experiments on the standard visual generation dataset ImageNet suggest that, with the powerful tokenizer, the plain auto-regressive model exhibits superiority and scalability. 
\end{itemize}

\subsection*{\textbf{Update Log}}
\begin{itemize}[leftmargin=0.5cm]
    \item \textbf{V0.1} (17/06/2024): Release an initial version of Open-MAGVIT2 image tokenizer.
    \item \textbf{V1.0} (09/09/2024): Release a series of Open-MAGVIT2 image tokenizers trained on ImageNet (dubbed as \texttt{Open-MAGVIT2-I-In1k} in the paper) and a family of auto-regressive models ranging from 300M to 1.5B (dubbed as \texttt{Open-MAGVIT2-AR-B/L/XL} in the paper).
    \item \textbf{V1.1} (21/01/2025): Release a pretrained version of Open-MAGVIT2 image tokenizers (dubbed as \texttt{Open-MAGVIT2-I-PT} in the paper).
    \item \textbf{V1.2} (09/02/2025): Release an Open-MAGVIT2 video tokenizer with 4$\times$ temporal downsampling rate (dubbed as \texttt{Open-MAGVIT2-V-UCF} in the paper).
\end{itemize}


\section{Method}
\subsection{Overview}
Open-MAGVIT2 is composed of two significant stages. One is a powerful visual tokenizer that maps the input visual signal into the discrete token representations. Subsequently, the vector-quantized sequence will be fed into the auto-regressive transformer for intra- and inter-token relationship modeling, eventually for visual synthesis.

\begin{figure*}
    \centering
    \includegraphics[width=\linewidth]{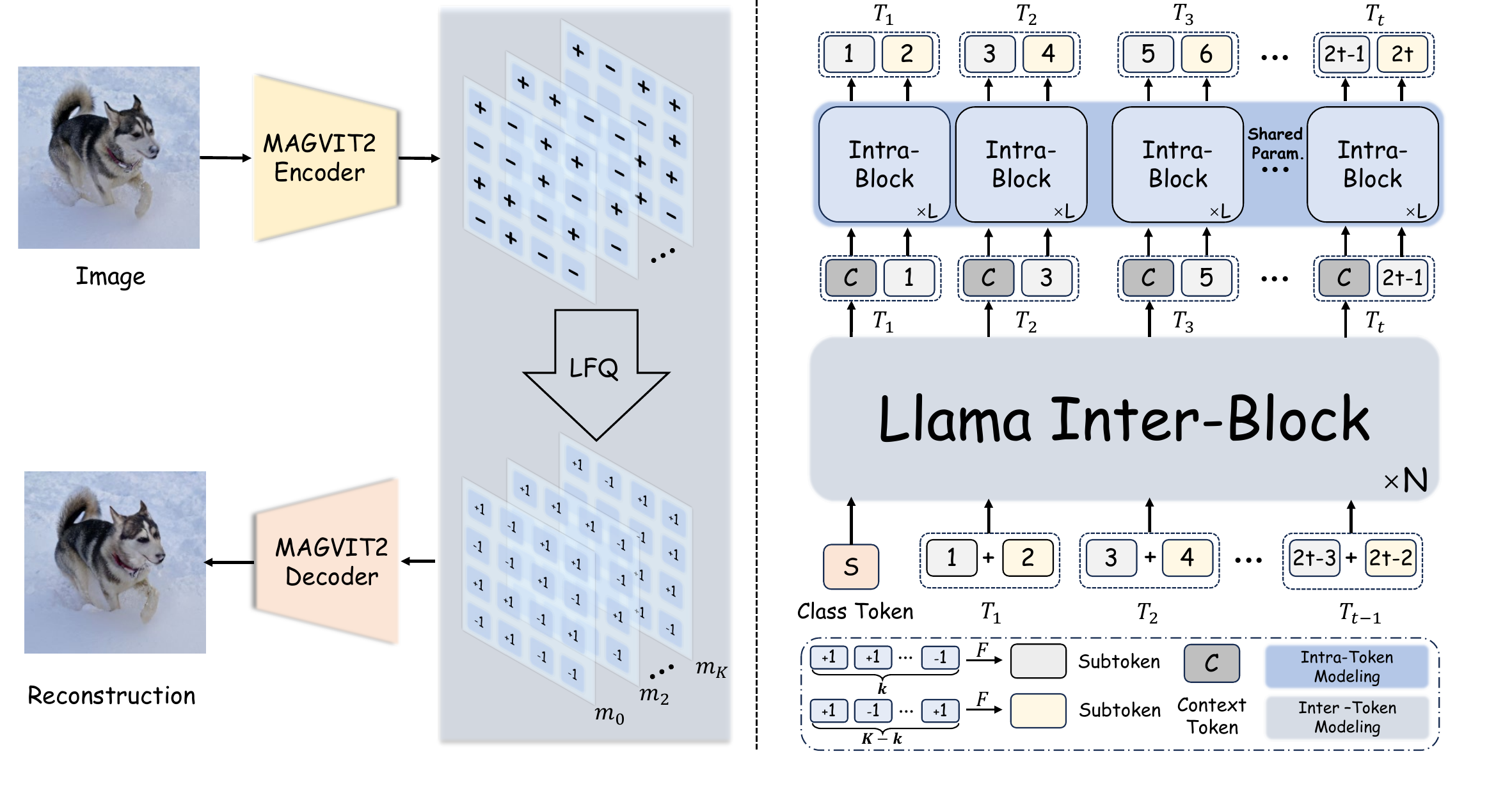}
    \vspace{-20pt}
    \caption{\textbf{Overview of Open-MAGVIT2.} There are two crucial stages in Open-MAGVIT2. In Stage $\mathrm{\uppercase\expandafter{\romannumeral1}}$: the image is first encoded by MAGVIT-v2 Encoder and subsequently transformed into bits format by Lookup-Free Quantizer (LFQ). In Stage $\mathrm{\uppercase\expandafter{\romannumeral2}}$: The quantized features are further mapped into discrete visual tokens and input into the Llama-based auto-regressive framework for intra- and inter-token relationship modeling.}
    \label{fig:framework}
    \vspace{-10pt}
\end{figure*}

\subsection{Visual Tokenizer}
\paragraph{Preliminary.} Visual tokenization is fundamentally deemed as the crucial component in multimodal large language models (MLLMs) to understand the visual signal input. The CNN-based encoder-quantizer-decoder architecture first proposed in VQVAE~\citep{vqvae} is well adopted as the visual tokenizer, which maps input pixels into discrete representations and reconstructs images from quantized features. Specifically, given an video $\mathcal{V} \in \mathbb{R}^{T \times 3 \times H \times W}$ (When $T=1$, the input is an image), the encoder projects it into the feature map $\mathcal{Z} \in \mathbb{R}^{T' \times D \times H' \times W'}$, where $T' = T / p_t, H' = H / p_s, W' = W / p_s,$ and $p_t$, $p_s$ are the temporal and spatial downsample ratio respectively. The quantizer containing a learnable codebook $\mathcal{E} \in \mathbb{R}^{2^{K} \times D}$ then selects the closest entry $\hat{z} \in \mathbb{R}^{D}$ from the codebook for each feature vector $z \in \mathbb{R}^{D}$. And we can use discrete token indices $\mathcal{X} = \{x_{i}\}_{i=1}^{T' \times H'\times W'}$ to represent the continuous feature map $\mathcal{Z}$. For decoding, each code index will be mapped back to the quantized feature vector and input into the decoder for pixel-level image reconstruction.

\begin{figure*}
    \centering
    \includegraphics[width=\linewidth]{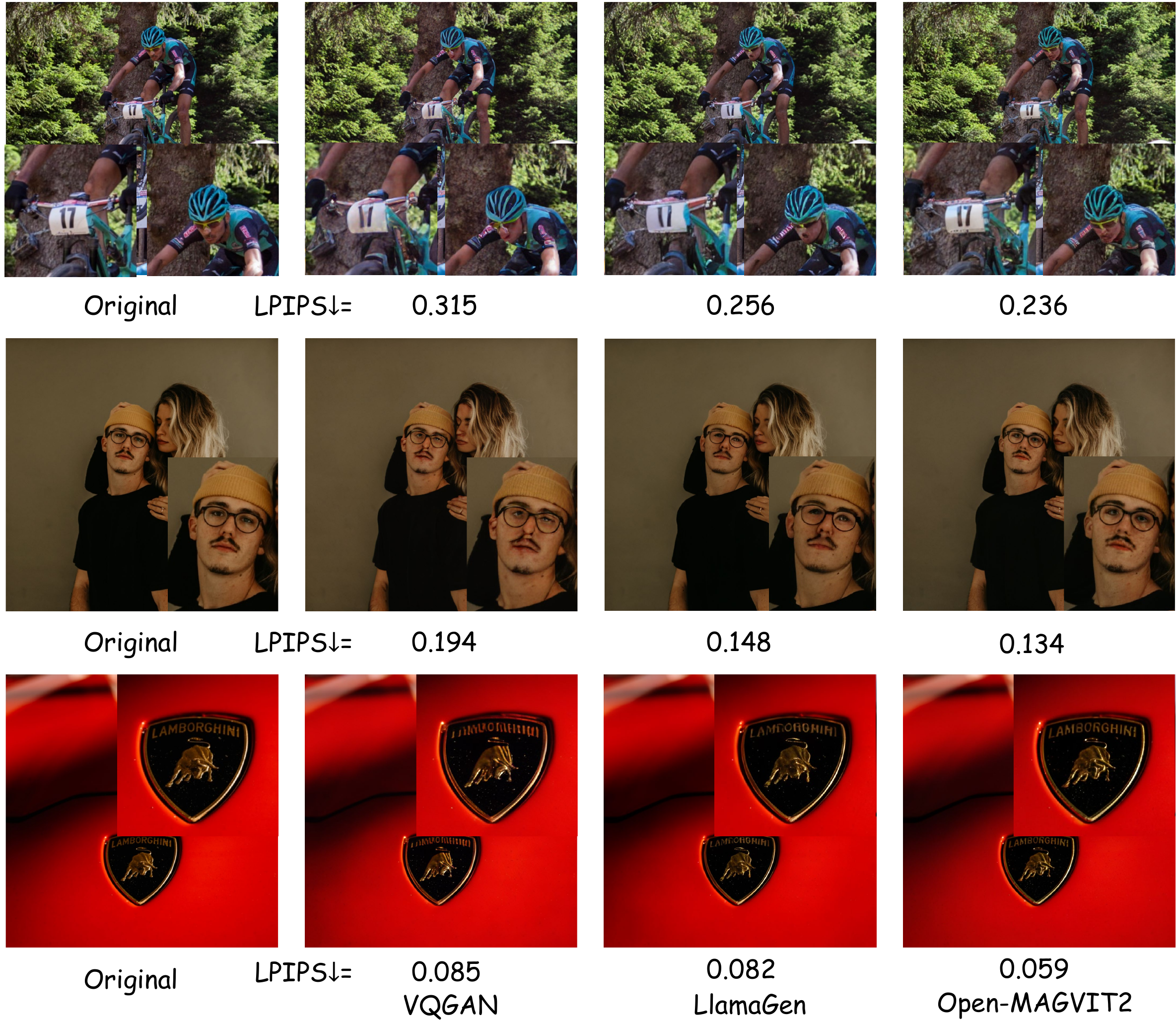}
    \vspace{-20pt}
    \caption{\textbf{Reconstruction comparison with different tokenizers.} We compare VQGAN~\citep{vqgan}, LlamaGen~\citep{llamagen} and our models trained on ImageNet. \textcolor{gray}{(Best viewed with zooming in. The original images are from Unsplash).}}
    \label{fig:teaser}
    \vspace{-10pt}
\end{figure*}

\begin{figure*}
    \centering
    \includegraphics[width=\linewidth]{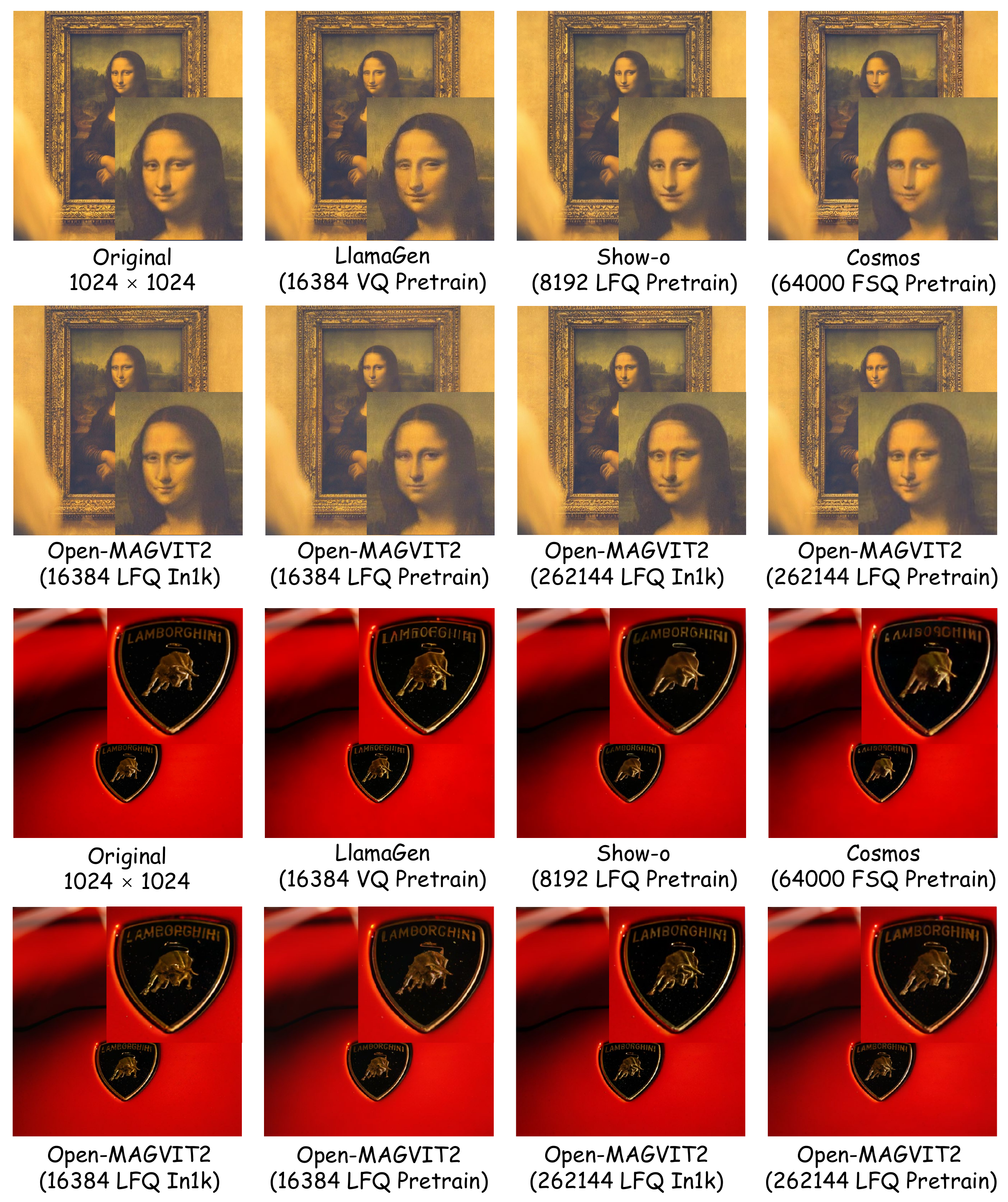}
    \vspace{-10pt}
    \caption{\textbf{Reconstruction comparison with different tokenizers .} We compare LlamaGen~\citep{llamagen}, Showo~\citep{showo}, Cosmos~\citep{cosmos} and our models. The bottom of each case illustrates that the tokenizer pretrained on large-scale datasets gains more superiority in facial and textual scenarios than ones trained on imagenet. Moreover, our pretrained visual tokenizer achieves better reconstruction soundness compared with LlamaGen, Show-o and Cosmos. \textcolor{gray}{(Best viewed with zooming in.)}}
    \label{fig:pretrain_comparsion}
    \vspace{-10pt}
\end{figure*}

\paragraph{Review of Lookup-Free Quantization.}
Motivated by the relationship between the size of the codebook and the dimension of code embeddings, MAGVIT-v2~\citep{magvit2} eliminates the need for embedding lookup by reducing the dimension of code embedding to zero. Specifically, the codebook is shrunk into an integer set where the latent space of each entry is decomposed as the Cartesian product of single-dimensional variables (i.e., $\hat{\mathcal{C}} = \bigtimes_{i=1}^{K} \{-1, 1\}, |\hat{\mathcal{C}}| = 2^K$). As shown in \cref{fig:framework}, the tokenization process can be simplified as:
\begin{equation}
    \hat{z}_i = \mathrm{sign}(z_i) =  -\mathbbm{1}\{z_i \leq 0\} + \mathbbm{1}\{z_i > 0\},
\end{equation}
where $\hat{z}_i$ denotes the quantized representation of the feature vector $z_i$. And the token index for $z_i$ is given by:
\begin{equation}
    Index(z_i) = \sum_{k=1}^{K}2^{k-1} \mathbbm{1} \{\hat{z}_{ik} > 0\}.
\end{equation}
To encourage the confident assignment of each codebook entry and utilization of the whole codebook simultaneously, MAGVIT-v2 further introduces entropy loss:
\begin{equation}
    \mathcal{L}_{entropy} = \frac{1}{BH'W'}\sum\mathbbm{H}\left(f\left(z\right)\right) - \mathbbm{H}(\frac{1}{BH'W'}\sum f\left(z\right)),
\end{equation}
where $\mathbbm{H}(\cdot)$ denotes the entropy, $B$ is batch sizes, and $f(\cdot)$ is a mapping function from latent space to a categorical distribution specifying the probability of assignment to each entry. In our experiment, we observe that replacing traditional code assignment (i.e., pair-wise distance) with this lookup-free quantization enables training a super-large codebook (i.e., $2^{18}$ codes) of high utilization ($100\%$).

\paragraph{Review of Architecture improvements.}
Intuitively, since each continuous feature vector will be quantized into $K$ bits, it poses a significant challenge to both the encoder and decoder. Therefore, we re-implement the architecture improvements technique illustrated in~\citep{magvit2}. 1) Downsamplers in the encoder are strided convolutions with learned kernels while upsamplers in the decoder are the depth-to-space operator. 2) Following~\citep{stylegan, dit, adaptin}, we re-implement the Adaptive GroupNorm Layer, which integrates the quantized vector with the output of each residual block in the decoder.   

\subsection{Auto-Regressive Transformer}
\paragraph{Preliminary.} Given a sequence of discrete tokens $\mathcal{X} = \{x_{i}\}_{i=1}^{T}, T=H' \times W'$ from the visual tokenizer, the auto-regressive transformer predicts the next token $x_{t}$ conditioned on the previous tokens $\{x_1, x_2, \cdots, x_{t-1}\}$:
\begin{equation}
p(x_1, x_2, \cdots, x_{T}) = \prod_{t=1}^{T}p(x_t|x_1, x_2, \cdots, x_{t-1}).
\end{equation}

\paragraph{Auto-regressive Architecture.}
Considering the different scales of auto-regressive transformer (i.e., from $\sim$300M to 1B) and the limited training academic data, directly optimizing such a large vocabulary (i.e., $2^{18}$ codes) is impractical. Therefore, we propose the asymmetric token factorization technique to assist models in performing ``next-token prediction'' within concatenated codebooks. Specifically, the LFQ token's latent space is factorized into $M$ subspaces $\{x_i^{1}\}_{i=1}^T$, $\{x_i^{2}\}_{i=1}^T$, $\cdots$, $\{x_i^{M}\}_{i=1}^T$, each of which contains $2^{k_m}$ tokens. As shown in \cref{fig:framework}, each subspace is embedded individually and their summation is used as the transformer inputs. 
Conventionally, an intuitive solution to perform auto-regressive within subspaces is leveraging $M$ separate heads for independent categorical distribution modeling. 
However, since both sub-tokens are derived from the same latent spaces, such a simple operation may ignore their intra-correlation.
Consequently, inspired by~\citep{rqvae}, we reformulate the autoregression paradigm into modeling both intra- and inter-token dependency, which is essentially ``next sub-token prediction''. In this manner, the representational capacity of the super-large codebook can exhibit great potential in auto-regressive generation with better scalability. 

1) \textbf{\textit{Inter-token Relationship}}: Given a set of sub-tokens from the visual tokenizers, a stacked of Llama blocks with $N$ layers and $w$ width are leveraged to capture the in-context information between tokens. The process can be formulated as:
\begin{equation}
    \mathcal{C}_t = \mathrm{LlamaBlock}(\mathit{s}, (\sum_{i=1}^{M}x_1^{i}), \cdots, (\sum_{i=1}^{M}x_{t-1}^{i})),
\end{equation}
where $\mathit{s}$ denotes the conditional tokens, $\mathcal{C}_{t} \in \mathbb{R}^{T \times w_s}$ is the $t$-th context token. 

2) \textbf{\textit{Intra-token Relationship}}: We further utilize a transformer with $L$ intra-blocks to autoregressively predict the each sub-token ($x_t^{1}$, $x_t^{2}, \cdots, x_t^{M}$) at the position $t$. By associating the sub-token conditioned with contextual-enriched vector $\mathcal{C}$, the intra-dependency within tokens can be well modeled. Formally, at $t$ position, the autoregression of predicting the conditional distribution of each sub-token is:
\begin{equation}
p_{tm} = \mathrm{LlamaBlock}(\mathcal{C}_t, x_t^{1} \cdots ,x_t^{m-1}).
\end{equation}
Therefore, the auto-regressive likelihood is formulated as:
\begin{equation}
\begin{aligned}
p(X_1, X_2, \cdots, X_T) &= \prod_{t=1}^{T} p(X_t|X_1, X_2, \cdots, X_{t-1}) \\
                        &= \prod_{t=1}^{T} \prod_{m=1}^{M}p(x_t^m|(X_1, X_2, \cdots, X_{t-1}), (x_t^{1}, x_t^{2}, \cdots x_t^{m-1})),   
\end{aligned}
\end{equation}
where $X_t$ specifies a set of sub-token $\{x_t^{1}, x_t^{2}, \cdots, x_t^{M}\}$ at each position t.

\section{Experiments}

\subsection{Dataset and Metrics}
\paragraph{Class-Conditional Image Generation.}
The training of the visual tokenizer and auto-regressive transformer are both on ImageNet~\citep{imagenet}. Specifically, we train the tokenizer in $128 \times 128$ and $256 \times 256$ resolutions.
For visual reconstruction, the reconstruction-FID, denoted as rFID~\citep{fid}, codebook utilization, the use percentage of codes, and PSNR on ImageNet 50k validation set are adopted to measure the quality of reconstructed images. Simultaneously, we measure the quality of image generation by the prevalent metrics FID, IS~\citep{is} and Precision/Recall~\citep{precision_recall}.
\vspace{-10pt}
\paragraph{Text-Conditional Image Generation.}
Aiming to serve text-conditional image generation, we pretrain the tokenizer on large-scale general-domain datasets, i.e., 1) General: LAION-COCO~\citep{laion_coco}, CC12M~\citep{cc12m} and CC3M~\citep{cc3m}. 2) High-quality: LAION-aesthetics-12M\footnote{https://huggingface.co/datasets/dclure/laion-aesthetics-12m-umap}, LAION-aesthetics~\citep{laion_aesthetics}, JourneyDB~\citep{journeydb} and LAION-HD\footnote{https://huggingface.co/datasets/yuvalkirstain/laion-hd-subset}. Following~\citep{cosmos}, we evaluate our visual tokenizer by the zero-shot performance on MS-COCO~\citep{coco} and Imagenet with rFID, PSNR, SSIM.
\paragraph{Video Generation.}
Following~\citep{magvit2}, we train our video tokenizer on UCF-101~\citep{ucf101}. We use 17-frame video clips with a spatial resolution of 128 $\times$ 128 for both training and evaluation. The reconstruction-FVD, denoted as rFVD~\citep{fvd} is adopted as the main metrics for quantifying the soundness of reconstructed videos.

\subsection{Implementations Details}
\paragraph{Visual Tokenizer Setup.}
Open-MAGVIT2 follows the same architecture of the visual tokenizer proposed in~\citep{magvit2}. The visual tokenizer for different purposes, e.g., class-conditional and text-conditional image generation, is trained with similar settings. Specifically, for computational efficiency, we remove the gradient penalty loss, and adopt PatchGAN~\citep{patchgan} as the discriminator instead of StyleGAN~\citep{stylegan}. All models corresponding to different resolutions are trained with similar settings: an initial $1e-4$ learning rate, an Adam Optimizer with $\beta_{1} = 0.5$, $\beta_2 = 0.9$, a total $256$ batch size from $270$ to $350$ epochs (1500k steps for text-conditional image generation), a combination of reconstruction, GAN, perceptual~\citep{lpips}, entropy penalty~\citep{magvit2}, commitment losses, LeCAM regularization~\citep{lecam} for training stability, and $32$ $\times$ GPU / NPU with Pytorch.

\begin{table}[t]
    \vspace{-5pt}
    \centering
    \setlength{\tabcolsep}{4pt}
    \renewcommand\arraystretch{1.0}
    \caption{Model designs and reconstruction performance comparison with the original MAGVIT-v2 on $128 \times 128$ ImageNet 50k validation set (rFID) and UCF-101 (rFVD). }
    \resizebox{\linewidth}{!}{
        \begin{tabular}{cccccccccc}
        \toprule
        \multirow{2}{*}{\textbf{Method}} & \multirow{2}{*}{\textbf{Tokens}} & \textbf{Train} & \multirow{2}{*}{\textbf{LFQ}} & \textbf{Large} & \textbf{Up/Down} & \textbf{Deeper} & \textbf{Adaptive} & \multirow{2}{*}{rFID} & \multirow{2}{*}{rFVD} \\
        & & \textbf{Resolution} & & \textbf{Codebook} & \textbf{Sampler} & \textbf{Model} & \textbf{GroupNorm} & \\
        \midrule
        Open-MAGVIT2 & $16 \times 16$ & $128 \times 128$ & $\checkmark$ & $\checkmark$ & $\checkmark$ & $\checkmark$ & $\checkmark$ & 1.18 & 16.0 \\
        MAGVIT2~\citep{magvit2} & $16 \times 16$ & $128 \times 128$ & $\checkmark$ & $\checkmark$ & $\checkmark$ & $\checkmark$ & $\checkmark$ & 1.15 & 8.6 \\
        \bottomrule
        \end{tabular}}
        \label{tab:recon2}
\end{table}

\begin{table}[t]
    \centering
    \setlength{\tabcolsep}{4pt}
    \renewcommand\arraystretch{1.1}
    \caption{Reconstruction performance of different tokenizers on
 ImageNet 50k validation set and UCF-101 dataset. For image tokenizers, Open-MAGVIT2 achieves SOTA results on different downsampling rates. $\dagger$ specifies that the training is on OpenImages. $*$ denotes that the results are from the direct inference using the model trained with $128 \times 128$ resolution without fine-tuning. For video tokenizers, Open-MAGVIT2 achieves the SOTA performance on the reconstruction results. $\dagger\dagger$ denotes that training SweetTokenizer without token compression.}
    \resizebox{\linewidth}{!}{
        \begin{tabular}{lcccccccc}
        \toprule
        \multirow{2}{*}{\textbf{Method}} & \textbf{Token} & \multirow{2}{*}{\textbf{Tokens}} & \multirow{2}{*}{\textbf{Ratio}} & \textbf{Train} & \textbf{Codebook} & \multirow{2}{*}{\textbf{rFI(V)D}$\downarrow$} & \multirow{2}{*}{\textbf{PSNR}$\uparrow$} & \textbf{Codebook} \\
        & \textbf{Type} &  & & \textbf{Resolution} & \textbf{Size} & & & \textbf{Usage}$\uparrow$ \\
        \midrule
        \multicolumn{9}{c}{\textit{Image Tokenizer}} \\
        \midrule
        VQGAN~\citep{vqgan} & 2D & $16 \times 16$ & 16 & $256 \times 256$ & 1024 & 7.94 & 19.4 & $-$ \\
        SD-VQGAN~\citep{ldm} & 2D & $16 \times 16$ & 16 & $256 \times 256$ & 16384 & 5.15 & $-$ & $-$ \\
        MaskGIT~\citep{maskgit} & 2D & $16 \times 16$ & 16 & $256 \times 256$ & 1024 & 2.28 & $-$ & $-$ \\
        LlamaGen~\citep{llamagen} & 2D & $16 \times 16$ & 16 & $256 \times 256$ & 16384 & 2.19 & 20.79 & 97\% \\
        \textbf{Open-MAGVIT2-I-In1k} & 2D & $16 \times 16$ & 16 &$256 \times 256$ & 262144 & \textbf{1.17} & \textbf{22.64} & \textbf{100\%} \\ \hline
        ViT-VQGAN~\citep{vit-vqgan} & 2D & $32 \times 32$ & 8 & $256 \times 256$ & 8192 & 1.28 & $-$ & $-$ \\
        VQGAN$^{\dagger}$~\citep{vqgan} & 2D & $32 \times 32$ & 8 & $256 \times 256$ & 16384 & 1.19 & 23.38 & $-$ \\
        SD-VQGAN$^{\dagger}$~\citep{ldm} & 2D & $32 \times 32$ & 8 & $256 \times 256$ &  16384 & 1.14 & $-$ & $-$ \\
        OmiTokenizer-VQ~\citep{omnitokenizer} & 2D & $32 \times 32$ & 8 & $256 \times 256$ & 8192 & 1.11 & $-$ & $-$ \\
        LlamaGen~\citep{llamagen} & 2D & $32 \times 32$ & 8 & $256 \times 256$ & 16384 & 0.59 & 24.45 & $-$ \\
        \textbf{Open-MAGVIT2-I-In1k} & 2D & $32 \times 32$ & 8 &$128 \times 128$ & 262144 & \textbf{0.34} & \textbf{27.02} & \textbf{100\%} \\ \hline
        Titok-L~\citep{titok} & 1D & 32 & $-$ & $256 \times 256$ & 4096 & 2.21 & $-$ & $-$ \\
        Titok-B~\citep{titok} & 1D & 64 & $-$ & $256 \times 256$ & 4096 & 1.70 & $-$ & $-$ \\
        Titok-S~\citep{titok} & 1D & 128 & $-$ & $256 \times 256$ & 4096 & 1.71 & $-$ & $-$ \\
        \midrule
        \multicolumn{9}{c}{\textit{Video Tokenizer}} \\
        \midrule
        TATS~\citep{tats} & 2D & $4 \times 16 \times 16$ & 8 & $128 \times 128$ & 16384 & 162 & $-$ & $-$ \\
        MAGVIT~\citep{magvit1} & 2D & $4 \times 16 \times 16$ & 8 & $128 \times 128$ & 1024 & 25 & $-$ & $-$ \\
        SweetTokenizer~\citep{sweettokenizer} & 1D & 256 + 1024 & $-$ & $256 \times 256$ & 10481 + 11139 & 44 & $-$ & $-$ \\
        LARP-L~\citep{larp}  & 1D & 1024 & $-$ & $128 \times 128$ &8192 & 24 & $-$ & $-$ \\
        LARP-L-Long~\citep{larp}  & 1D & 1024 & $-$ & $128 \times 128$ & 8192 & 20 & $-$ & $-$ \\
        SweetTokenizer$^{\dagger\dagger}$~\citep{sweettokenizer} & 1D & 5120 & $-$ & $256 \times 256$ & 10481 + 11139 & 18 & $-$ & $-$ \\
        \textbf{Open-MAGVIT2-V-UCF} & 2D & $5 \times 16 \times 16$ & 8 & $128 \times 128$ & 262144 & \textbf{16} & $-$ & 100 \\
    \bottomrule
    \end{tabular}}
    \label{tab:recon1}
\end{table}

\begin{table}[t]
    \centering
    \setlength{\tabcolsep}{4pt}
    \renewcommand\arraystretch{1.0}
    \caption{Class-conditional generation on $256 \times 256$ ImageNet. $*$ specifies the generated images are $384 \times 384$ and are resized to 256×256 for evaluation. The evaluation protocol and implementation are the same with ADM.}
    \resizebox{0.9\linewidth}{!}{
    \begin{tabular}{clccccc}
    \toprule
    \textbf{Type} & \textbf{Model} & \textbf{\#Para.} & \textbf{FID}$\downarrow$ & \textbf{IS}$\uparrow$ & \textbf{Precision}$\uparrow$ & \textbf{Recall}$\uparrow$  \\
    \midrule
    \multirow{4}{*}{\textcolor{gray}{Diffusion}} & \textcolor{gray}{ADM}~\citep{adm}  & \textcolor{gray}{554M}       & \textcolor{gray}{10.94} & \textcolor{gray}{101.0}        & \textcolor{gray}{0.69} & \textcolor{gray}{0.63}    \\
     & \textcolor{gray}{CDM}~\citep{cdm}   & \textcolor{gray}{$-$}       & \textcolor{gray}{4.88}  & \textcolor{gray}{158.7}       & \textcolor{gray}{$-$}  & \textcolor{gray}{$-$}   \\
     & \textcolor{gray}{LDM-4}~\citep{ldm} & \textcolor{gray}{400M}     & \textcolor{gray}{3.60}  & \textcolor{gray}{247.7}       & \textcolor{gray}{$-$}  & \textcolor{gray}{$-$}  \\
     & \textcolor{gray}{DiT-XL/2}~\citep{dit}  & \textcolor{gray}{675M}  & \textcolor{gray}{2.27}  & \textcolor{gray}{278.2}       & \textcolor{gray}{0.83} & \textcolor{gray}{0.57}   \\
    \midrule
    \multirow{7}{*}{AR} & VQGAN~\citep{vqgan} & 227M & 18.65 & 80.4         & 0.78 & 0.26    \\
     & VQGAN~\citep{vqgan}    & 1.4B   & 15.78 & 74.3   & $-$  & $-$     \\
     & VQGAN-re~\citep{vqgan}  & 1.4B  & 5.20  & 280.3  & $-$  & $-$     \\
     & ViT-VQGAN~\citep{vit-vqgan} & 1.7B & 4.17  & 175.1  & $-$  & $-$        \\
     & ViT-VQGAN-re~\citep{vit-vqgan}& 1.7B  & 3.04  & 227.4  & $-$  & $-$     \\
     & RQTran.~\citep{rqvae}       & 3.8B  & 7.55  & 134.0  & $-$  & $-$     \\
     & RQTran.-re~\citep{rqvae}    & 3.8B & 3.80  & 323.7  & $-$  & $-$    \\
    \midrule
    \multirow{4}{*}{VAR} & VAR-d16~\citep{var} & 310M & 3.30 & 274.4 & 0.84 & 0.51 \\
    & VAR-d20~\citep{var} & 600M & 2.57 & 302.6 & 0.83 & 0.56 \\
    & VAR-d24~\citep{var} & 1.0B & 2.09 & 312.9 & 0.82 & 0.59 \\
    & VAR-d30~\citep{var} & 2.0B & 1.92 & 323.1 & 0.82 & 0.59 \\
    \midrule
    \multirow{11}{*}{AR} & LlamaGen-L$^{*}$~\citep{llamagen} & 343M & 3.07 & 256.06 & 0.83 & 0.52 \\
    & LlamaGen-XL$^{*}$~\citep{llamagen} & 775M & 2.62 & 244.08 & 0.80 & 0.57 \\
    & LlamaGen-XXL$^{*}$~\citep{llamagen} & 1.4B & 2.34 & 253.90 & 0.80 & 0.59 \\
    & LlamaGen-L~\citep{llamagen} & 343M & 3.80 & 248.28 & 0.83 & 0.51 \\
    & LlamaGen-XL~\citep{llamagen} & 775M & 3.39 & 227.08 & 0.81 & 0.54 \\
    & LlamaGen-XXL~\citep{llamagen} & 1.4B & 3.09 & 253.61 & 0.83 & 0.53 \\
     & Open-MAGVIT2-AR-B & 343M & 3.08 & 258.26 & 0.85 & 0.51 \\
     & Open-MAGVIT2-AR-L & 804M & 2.51 & 271.70 & 0.84 & 0.54 \\
     & Open-MAGVIT2-AR-XL & 1.5B & 2.33 & 271.77 & 0.84 & 0.54 \\
    \bottomrule
    \end{tabular}}
    \label{tab:gen1}
\end{table}

\begin{table}[t]
    \centering
    \setlength{\tabcolsep}{4pt}
    \renewcommand\arraystretch{1.1}
    \caption{Zero-shot reconstruction performance on ImageNet 50k validation set and MS-COCO val2017. The tokenizers are trained with large-scale general-domain datasets and aim to serve text-conditional image generation. The results are reported under the same setup for fair comparison (\textcolor{gray}{text in gray signifies the results directly from Cosmos~\citep{cosmos} report}). 
    $\dagger$ indicates that LlamaGen loads the model initially trained on Imagenet while the others are training from scratch, i.e., MS-COCO and Imagenet-1k are excluded from training data.}
    \resizebox{\linewidth}{!}{
        \begin{tabular}{lccccccccccc}
        \toprule
        \multirow{2}{*}{\textbf{Method}} & \textbf{Quantizer} & \textbf{Training} & \multirow{2}{*}{\textbf{Ratio}} & \textbf{Codebook} & 
        \multicolumn{3}{c}{\textbf{MS-COCO 2017}} & & \multicolumn{3}{c}{\textbf{Imagenet-1k}} \\
        \cmidrule{6-8} \cmidrule{10-12}
        & \textbf{Type} & \textbf{Data} & & \textbf{Size} & \textbf{rFID}$\downarrow$ & \textbf{PSNR}$\uparrow$ & \textbf{SSIM}$\uparrow$ & & \textbf{rFID}$\downarrow$ & \textbf{PSNR}$\uparrow$ & \textbf{SSIM}$\uparrow$ \\
        \midrule
        \multicolumn{12}{c}{\textit{Resize $256 \times 256$}} \\
        \midrule
        LlamaGen$^{\dagger}$~\citep{llamagen} & VQ & 70M & 16 & 16384 & 8.40 & 20.28 & 0.55 & & 2.47 & 20.65 & 0.54 \\
        Show-o~\citep{showo} & LFQ & 35M & 16 & 8192 & 9.26 & 20.90 & 0.59 & & 3.50 & 21.34 & 0.59 \\
        Cosmos~\citep{cosmos} & FSQ & - & 16 & 64000 & 11.97 & 19.22 & 0.48 & & 4.57 & 19.93 & 0.49 \\ 
        Open-MAGVIT2-I-PT & LFQ & 100M & 16 & 16384 & 7.93 & 22.21 & 0.62 & & 2.55 & 22.21 & 0.62 \\
        \textbf{Open-MAGVIT2-I-PT} & LFQ & 100M & 16 & 262144 & 6.76 & 22.31 & 0.65 & & 1.67 & 22.70 & 0.64 \\
        \midrule
        \multicolumn{12}{c}{\textit{Original Resolution}} \\
        \midrule
        Cosmos~\citep{cosmos} & FSQ & - & 16 & 64000 & 7.51 & 20.45 & 0.52 & & 1.93 & 20.56 & 0.51 \\
        \textcolor{gray}{Cosmos}~\citep{cosmos} & \textcolor{gray}{FSQ} & - & \textcolor{gray}{16} & \textcolor{gray}{64000} & \textcolor{gray}{7.23} & \textcolor{gray}{20.45} & \textcolor{gray}{0.53} & & \textcolor{gray}{2.52} & \textcolor{gray}{20.49} & \textcolor{gray}{0.52} \\
        Open-MAGVIT2-I-PT & LFQ & 100M & 16 & 16384 & 6.65 & 21.61 & 0.57 & & 1.39 & 21.74 & 0.56 \\
        \textbf{Open-MAGVIT2-I-PT} & LFQ & 100M & 16 & 262144 & 5.10 & 22.18 & 0.60 & & 0.78 & 22.24 & 0.59 \\
    \bottomrule
    \end{tabular}}
    \label{tab:pretrain_recon}
\end{table}

\paragraph{Video Tokenizer Setup.}
We extend the image version of tokenizer following~\citep{magvit2} where the temporally casual 3D convolution serves as the main design. Simultaneously, we inflate the image tokenizer trained at 128 $\times$ 128 imagenet for video modeling. Similarly, we adopt 3D PatchGAN as the discriminator. The model is trained with the following settings: an initial $1e-4$ learning rate with a cosine annealing scheduler, an Adam Optimizer with $\beta_{1} = 0.5$, $\beta_2 = 0.9$, a total $128$ batch size with 2500 epochs, the losses are the same as the image tokenizer and $64$ $\times$ GPU / NPU with Pytorch.

\paragraph{Auto-regressive Transformer Setup.}
As illustrated, we propose asymmetric token factorization to assist the auto-regressive transformer models in making the precise prediction with a large codebook. Note that, we empirically set $M=2$ and $k_1=6$, $k_2=12$. 
Since our main focus is on democratizing scalable auto-regressive visual generation, the plain auto-regressive transformer is utilized while the techniques that introduce inductive bias such as AdaLn~\citep{adaln2} are excluded. Specifically, we adopt the Llama-based~\citep{llama2} architecture (e.g., RoPE~\citep{rope}, SwiGLU~\citep{glu}, RMSNorm~\citep{rmsnorm} technique, each of which has been proven effective in~\citep{llamagen}). The class embedding which is indexed from a set of learnable embeddings serves as the start token. Open-MAGVIT2 follows the simple scaling principle proposed in~\citep{llamagen}, which is in \cref{tab:scaling}.
All models are trained with similar settings: a base learning rate of $1e-4$ per $256$ batch size, an AdamW optimizer with $\beta_{1} = 0.9$, $\beta_2 = 0.95$, weight decay = $5e-2$, a total $768$ batch size and 300 $\sim$ 350 training epochs, gradient clipping of $1.0$, $0.1$ dropout rate for input embedding, FFN module and conditional embedding, $32$ $\sim$ 96 $\times$ GPU / NPU for different scales of the model with Pytorch. 

\subsection{Main Results}

\paragraph{Visual Reconstruction.} 
As shown in \cref{tab:recon2}, by incorporating all useful designs proposed in~\citep{magvit2}, Open-MAGVIT2 matches MAGVIT-v2 performances with merely $0.03$ FID margin on ImageNet $128 \times 128$. Further, we also compare our Open-MAGVIT2 with previous visual tokenizers on ImageNet $256 \times 256$ in \cref{tab:recon1}. Benefiting from the super-large codebook with lookup-free quantization, Open-MAGVIT2 outperforms all previous image tokenizers under fair settings. Moreover, we provide an illustrative visual comparison in \cref{fig:teaser}. As indicated, our visual tokenizer gains more superiority in detail perception as well as precise facial and text reconstruction. 

To facilitate the development of autoregressive text-conditional image generation, we provide an improved version of our tokenizers by pretraining on large-scale image-text datasets. As illustrated in \cref{fig:pretrain_comparsion}, our tokenizers achieve state-of-the-art performance compared to concurrent methods such as Cosmos~\citep{cosmos} in terms of zero-shot reconstruction on both ImageNet and COCO datasets. It is worth noting that some recent efforts in residual tokenization~\citep{tokenflow, infinity} can achieve better results, but are not listed here because residual techniques are orthogonal and compatible with quantization methods such as VQ, LFQ, and FSQ.

Furthermore, we extend our image tokenizer to the video version. As depicted in \cref{tab:recon2}, The reconstruction result gap on UCF-101 between MAGVIT-v2 and us is narrow. It is worth noting that Open-MAGVIT2 achieves a competitive performance among previous video tokenizers (see in \cref{tab:recon1})


\paragraph{Visual Generation.}
MAGVIT-v2 leverages the non-autoregressive framework for image synthesis and achieves competitive performance. Considering the scalability of auto-regressive models and the remarkable success of the auto-regressive paradigm in MLLM~\citep{chameleon}, we instead focus on exploring the potential of incorporating a super-large codebook for auto-regressive visual generation. As shown in \cref{tab:gen1}, Open-MAGVIT2 outperforms all previous image generation models using a plain auto-regressive approach. This benefits from the increased representational capacity of the large scale of the codebook. However, we believe that the strength of such a large codebook is still underestimated because of the data bottleneck and the model size. We hope our effort in building such a powerful visual tokenizer helps merit future research in unified MLLM for image generation. 

\subsection{Qualitative Results}
\begin{figure*}
    \centering
    \includegraphics[width=1.0\linewidth]{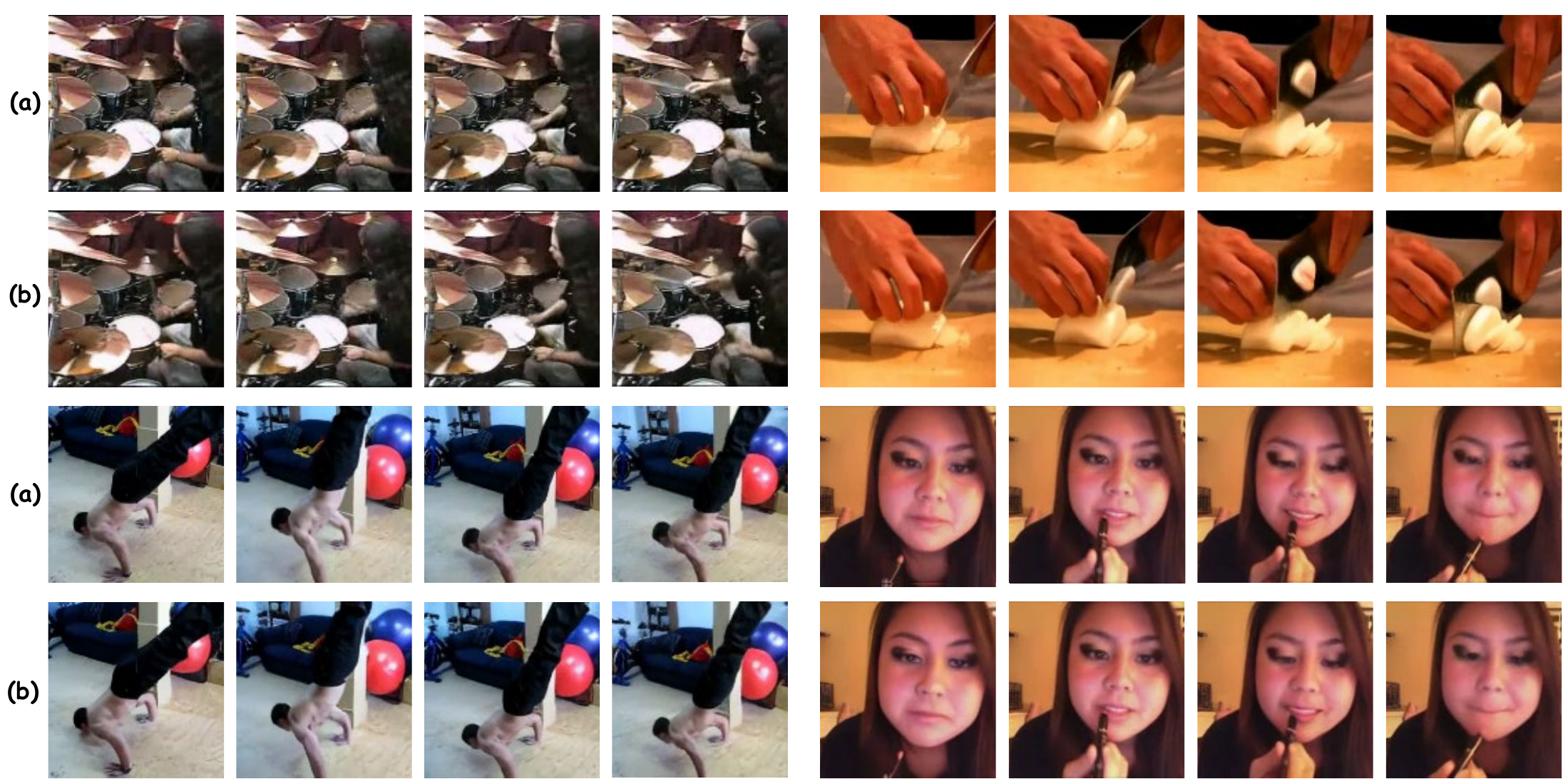}
    \vspace{-3pt}
    \caption{\textbf{Visualization of Open-MAGVIT2 video reconstruction}. (a) indicates the original videos while (b) specifies the reconstruction videos.}
    \label{fig:video_reconstruction}
    \vspace{-10pt}
\end{figure*}

\begin{figure*}
    \centering
    \includegraphics[width=1.0\linewidth]{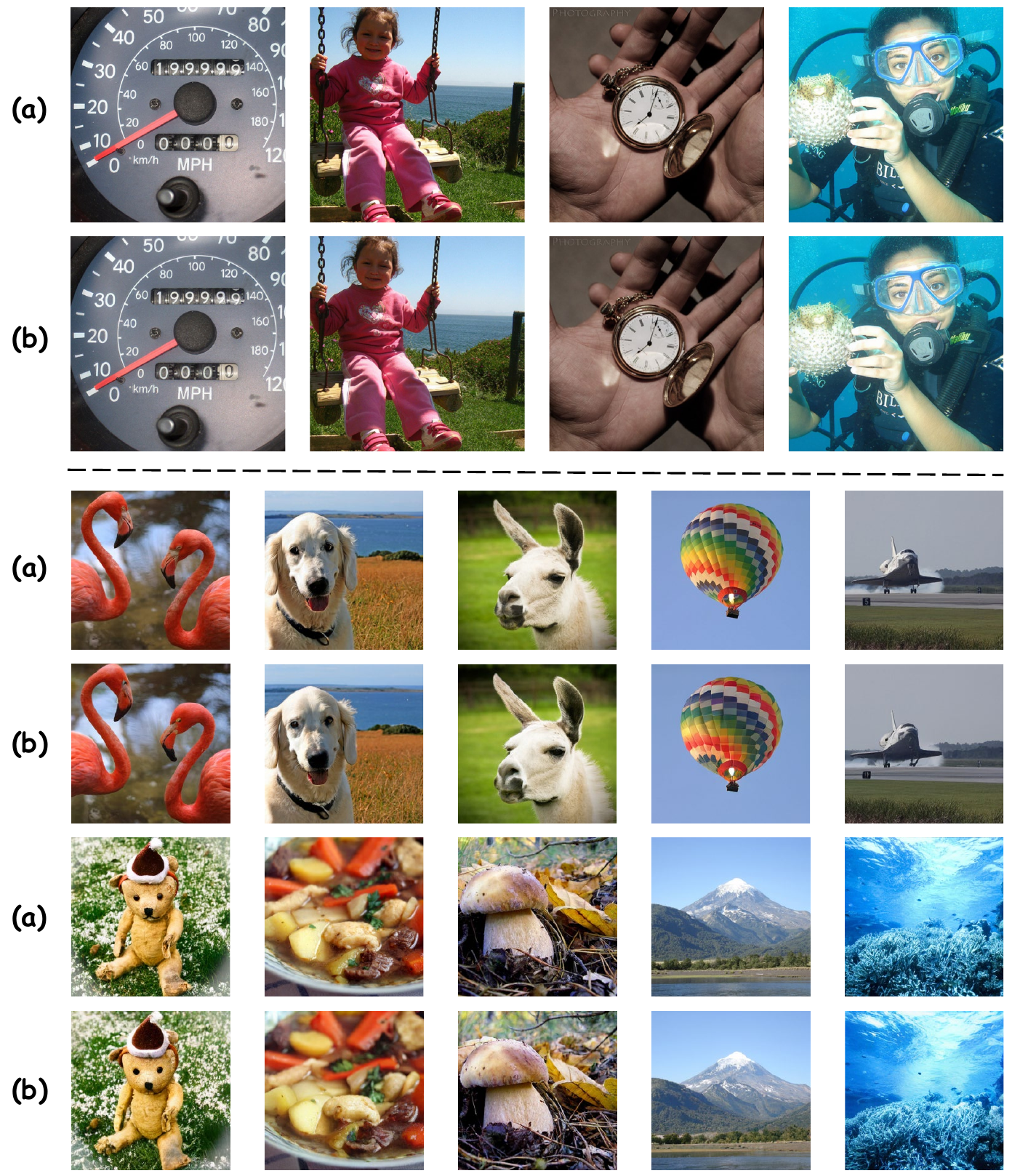}
    \vspace{-3pt}
    \caption{\textbf{Visualization of the Open-MAGVIT2 tokenizer}. The upper part illustrates the model trained at $128 \times 128$ resolution and tested at $512 \times 512$ resolution. The second part showcases the tokenizer trained at $256 \times 256$ resolution and tested at $256 \times 256$ resolution. (a) indicates the original images while (b) specifies the reconstruction images.}
    \label{fig:reconstruction}
    \vspace{-10pt}
\end{figure*}

\begin{figure*}
    \centering
    \includegraphics[width=1.0\linewidth]{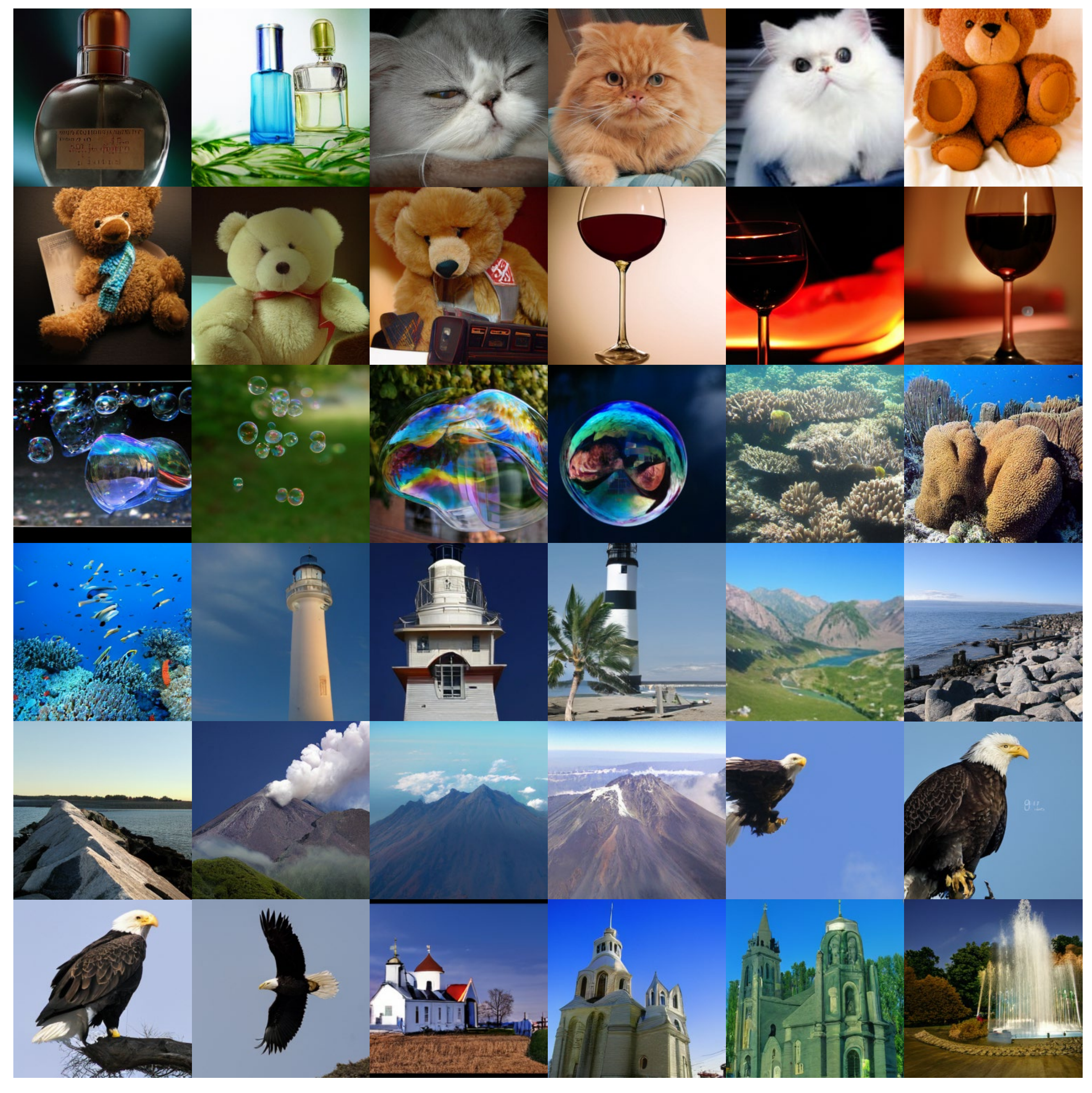}
    \vspace{-3pt}
    \caption{\textbf{Visualization of Open-MAGVIT2 auto-regressive generations}. Class-conditional generation on ImageNet $256\times 256$.}
    \label{fig:generation}
    \vspace{-10pt}
\end{figure*}

We present the qualitative results on Imagenet Benchmark in terms of visual reconstruction (see in \cref{fig:reconstruction}) and visual generation (see in \cref{fig:generation}), respectively. The video reconstruction visualization (see in \cref{fig:video_reconstruction}) are on UCF-101.

\section{Related Works}

\subsection{Visual Tokenizer}
Visual tokenizer is to map an image into compact discrete tokens, which are subsequently fed into the generative models for sequence modeling. Early pioneer VQVAE~\citep{vqvae} first introduces learnable codebook mechanism for 2D tokens generation. Subsequently, ViT-VQGAN~\citep{vit-vqgan} and RQ-VAE~\citep{rqvae} improve VQVAE through normalized and multi-scale quantization respectively. Recently, LlamaGen~\citep{llamagen} reexamines the design of vanilla tokenizer~\citep{vqgan} and reveals the conflict between the fidelity of the synthesized image and the size of codebook. Therefore, following the simple intuition~\citep{vit-vqgan} that reducing code dimension limits the representational capacity of individual tokens, MAGVIT-2~\citep{magvit2} proposes an advanced visual tokenizer which significantly enlarges the size of codebook to $2^{18}$ with Lookup-Free Quantization. 

\subsection{Visual Generation}
Given a set of compact discrete image tokens, there exist two prevalent frameworks for the subsequent image synthesis, including Non-autoregressive and Auto-regressive generation. 
\paragraph{Non-autoregressive frameworks.} MaskGIT~\citep{maskgit} utilizes BERT-style transformer~\citep{bert} to parallelly generate all visual tokens via masked-prediction mechanism. MAGVIT~\citep{magvit1, magvit2} adopts the same architecture but includes an additional embedding mask for better generation quality. 

\paragraph{Auto-regressive frameworks.}
Autoregressive-based Multi-Modal Large Language Models~\citep{llava, mini-gemini} has achieved remarkable success in versatile visual understanding. In contrast, the progress in counterpart 
visual generation still remains unsatisfactory. The simplest approach VQGAN~\citep{vqgan} employs tiny GPT2~\citep{gpt2} ($\sim$ 300M) for next-token prediction. VAR~\citep{var} reformulates the image generation approach into next-scale prediction and unveils the scaling principle simultaneously. Subsequently, LlamaGen~\citep{llamagen} extends VQGAN with Llama~\citep{llama2} architecture, showcasing significant improvement in fidelity. However, the limited codebook size (e.g., $2^{14}$) in existing auto-regressive models may incur the representational bottleneck. Therefore, considering that the capacity of the visual tokenizer is highly correlated with the quality of visual synthesis~\citep{magvit2}, we democratize the plain auto-regressive approach with a super-large codebook.


\section{Conclusion}
In this work, we re-implement the powerful visual tokenizer, which achieves state-of-the-art performance compared with previous methods, and make it available to the community. Instead of simply following~\citep{magvit2} that leverages masked-generative transformer for visual generation, we delve into a more promising manner (i.e., auto-regressive visual synthesis). To excavate the potential of the large vocabulary, we introduce the ``next sub-token prediction'' paradigm with the 
asymmetric token factorization technique. The experiment suggests that with the powerful tokenizer, the plain
auto-regressive model exhibits superiority and scalability. We hope our contribution to the open-source community can facilitate more innovative and creative works in the field of auto-regressive visual generation, eventually making a difference in building an omnipotent multi-modal framework.

\paragraph{Limitations and future work.}
We expect that the effectiveness of such a super-large codebook, (i.e., $2^{18}$ codes), is still underestimated due to the limited data scale and the sacrifice of the representational capacity with the token factorization technique. 
We believe that by amplifying the task with more training data (e.g., text-conditional image generation, video generation, etc.), and enlarging the model size to $7$B or even larger, the potential of AR generation with a super-large codebook can be dramatically exploited. Therefore, extending Open-MAGVIT2 into more broad multi-modal generation applications will be a high priority in our future exploration.

\subsubsection*{Acknowledgments}
We sincerely thank Lijun Yu for his encouraging discussions and support. We also thank Tianheng Cheng and Yuxin Chen for their helpful suggestions on this project.

\clearpage
\newpage
\bibliography{iclr2024_conference}

\begin{thebibliography}{54}
\providecommand{\natexlab}[1]{#1}
\providecommand{\url}[1]{\texttt{#1}}
\expandafter\ifx\csname urlstyle\endcsname\relax
  \providecommand{\doi}[1]{doi: #1}\else
  \providecommand{\doi}{doi: \begingroup \urlstyle{rm}\Url}\fi

\bibitem[Agarwal et~al.(2025)Agarwal, Ali, Bala, Balaji, Barker, Cai, Chattopadhyay, Chen, Cui, Ding, et~al.]{cosmos}
Niket Agarwal, Arslan Ali, Maciej Bala, Yogesh Balaji, Erik Barker, Tiffany Cai, Prithvijit Chattopadhyay, Yongxin Chen, Yin Cui, Yifan Ding, et~al.
\newblock Cosmos world foundation model platform for physical ai.
\newblock \emph{arXiv preprint arXiv:2501.03575}, 2025.

\bibitem[Chang et~al.(2022)Chang, Zhang, Jiang, Liu, and Freeman]{maskgit}
Huiwen Chang, Han Zhang, Lu~Jiang, Ce~Liu, and William~T. Freeman.
\newblock Maskgit: Masked generative image transformer.
\newblock In \emph{CVPR}, pp.\  11305--11315, 2022.

\bibitem[Changpinyo et~al.(2021)Changpinyo, Sharma, Ding, and Soricut]{cc12m}
Soravit Changpinyo, Piyush Sharma, Nan Ding, and Radu Soricut.
\newblock Conceptual 12m: Pushing web-scale image-text pre-training to recognize long-tail visual concepts.
\newblock In \emph{CVPR}, pp.\  3558--3568, 2021.

\bibitem[Chowdhery et~al.(2022)Chowdhery, Narang, Devlin, Bosma, Mishra, Roberts, Barham, Chung, Sutton, Gehrmann, et~al.]{palm}
Aakanksha Chowdhery, Sharan Narang, Jacob Devlin, Maarten Bosma, Gaurav Mishra, Adam Roberts, Paul Barham, Hyung~Won Chung, Charles Sutton, Sebastian Gehrmann, et~al.
\newblock Pa{LM}: Scaling language modeling with pathways.
\newblock \emph{arXiv preprint arXiv:2204.02311}, 2022.

\bibitem[Deng et~al.(2009)Deng, Dong, Socher, Li, Li, and Fei-Fei]{imagenet}
Jia Deng, Wei Dong, Richard Socher, Li-Jia Li, Kai Li, and Li~Fei-Fei.
\newblock Image{N}et: A large-scale hierarchical image database.
\newblock In \emph{CVPR}, pp.\  248--255, 2009.

\bibitem[Devlin et~al.(2018)Devlin, Chang, Lee, and Toutanova]{bert}
Jacob Devlin, Ming-Wei Chang, Kenton Lee, and Kristina Toutanova.
\newblock Bert: Pre-training of deep bidirectional transformers for language understanding.
\newblock \emph{arXiv preprint arXiv:1810.04805}, 2018.

\bibitem[Dhariwal \& Nichol(2021)Dhariwal and Nichol]{adm}
Prafulla Dhariwal and Alexander Nichol.
\newblock Diffusion models beat gans on image synthesis.
\newblock \emph{NeurIPS}, 34:\penalty0 8780--8794, 2021.

\bibitem[Esser et~al.(2021)Esser, Rombach, and Ommer]{vqgan}
Patrick Esser, Robin Rombach, and Bj{\"{o}}rn Ommer.
\newblock Taming transformers for high-resolution image synthesis.
\newblock In \emph{CVPR}, pp.\  12873--12883, 2021.

\bibitem[Ge et~al.(2022)Ge, Hayes, Yang, Yin, Pang, Jacobs, Huang, and Parikh]{tats}
Songwei Ge, Thomas Hayes, Harry Yang, Xi~Yin, Guan Pang, David Jacobs, Jia-Bin Huang, and Devi Parikh.
\newblock Long video generation with time-agnostic vqgan and time-sensitive transformer.
\newblock In \emph{ECCV}, pp.\  102--118, 2022.

\bibitem[Han et~al.(2024)Han, Liu, Jiang, Yan, Zhang, Yuan, Peng, and Liu]{infinity}
Jian Han, Jinlai Liu, Yi~Jiang, Bin Yan, Yuqi Zhang, Zehuan Yuan, Bingyue Peng, and Xiaobing Liu.
\newblock Infinity: Scaling bitwise autoregressive modeling for high-resolution image synthesis.
\newblock \emph{arXiv preprint arXiv:2412.04431}, 2024.

\bibitem[Heusel et~al.(2017)Heusel, Ramsauer, Unterthiner, Nessler, and Hochreiter]{fid}
Martin Heusel, Hubert Ramsauer, Thomas Unterthiner, Bernhard Nessler, and Sepp Hochreiter.
\newblock {GAN}s trained by a two time-scale update rule converge to a local nash equilibrium.
\newblock In \emph{NeurIPS}, volume~30, 2017.

\bibitem[Ho et~al.(2022)Ho, Saharia, Chan, Fleet, Norouzi, and Salimans]{cdm}
Jonathan Ho, Chitwan Saharia, William Chan, David~J Fleet, Mohammad Norouzi, and Tim Salimans.
\newblock Cascaded diffusion models for high fidelity image generation.
\newblock \emph{JMLR}, 23\penalty0 (1):\penalty0 2249--2281, 2022.

\bibitem[Huang \& Belongie(2017)Huang and Belongie]{adaptin}
Xun Huang and Serge~J. Belongie.
\newblock Arbitrary style transfer in real-time with adaptive instance normalization.
\newblock In \emph{ICCV}, pp.\  1510--1519, 2017.

\bibitem[Isola et~al.(2017)Isola, Zhu, Zhou, and Efros]{patchgan}
Phillip Isola, Jun{-}Yan Zhu, Tinghui Zhou, and Alexei~A. Efros.
\newblock Image-to-image translation with conditional adversarial networks.
\newblock In \emph{CVPR}, pp.\  5967--5976, 2017.

\bibitem[Karras et~al.(2019)Karras, Laine, and Aila]{stylegan}
Tero Karras, Samuli Laine, and Timo Aila.
\newblock A style-based generator architecture for generative adversarial networks.
\newblock In \emph{CVPR}, pp.\  4401--4410, 2019.

\bibitem[Karras et~al.(2020)Karras, Laine, Aittala, Hellsten, Lehtinen, and Aila]{adaln2}
Tero Karras, Samuli Laine, Miika Aittala, Janne Hellsten, Jaakko Lehtinen, and Timo Aila.
\newblock Analyzing and improving the image quality of stylegan.
\newblock In \emph{CVPR}, pp.\  8110--8119, 2020.

\bibitem[Kingma(2013)]{vae}
Diederik~P Kingma.
\newblock Auto-encoding variational bayes.
\newblock \emph{arXiv preprint arXiv:1312.6114}, 2013.

\bibitem[Kynk{\"{a}}{\"{a}}nniemi et~al.(2019)Kynk{\"{a}}{\"{a}}nniemi, Karras, Laine, Lehtinen, and Aila]{precision_recall}
Tuomas Kynk{\"{a}}{\"{a}}nniemi, Tero Karras, Samuli Laine, Jaakko Lehtinen, and Timo Aila.
\newblock Improved precision and recall metric for assessing generative models.
\newblock In Hanna~M. Wallach, Hugo Larochelle, Alina Beygelzimer, Florence d'Alch{\'{e}}{-}Buc, Emily~B. Fox, and Roman Garnett (eds.), \emph{NeurIPS}, pp.\  3929--3938, 2019.

\bibitem[LAION(2022)]{laion_coco}
LAION.
\newblock Laion-coco 600m.
\newblock \url{https://laion.ai/blog/laion-coco}, 2022.

\bibitem[Lee et~al.(2022)Lee, Kim, Kim, Cho, and Han]{rqvae}
Doyup Lee, Chiheon Kim, Saehoon Kim, Minsu Cho, and Wook{-}Shin Han.
\newblock Autoregressive image generation using residual quantization.
\newblock In \emph{CVPR}, pp.\  11513--11522, 2022.

\bibitem[Li et~al.(2024)Li, Zhang, Wang, Zhong, Chen, Chu, Liu, and Jia]{mini-gemini}
Yanwei Li, Yuechen Zhang, Chengyao Wang, Zhisheng Zhong, Yixin Chen, Ruihang Chu, Shaoteng Liu, and Jiaya Jia.
\newblock Mini-gemini: Mining the potential of multi-modality vision language models.
\newblock \emph{arXiv preprint arXiv:2403.18814}, 2024.

\bibitem[Lin et~al.(2014)Lin, Maire, Belongie, Hays, Perona, Ramanan, Doll{\'a}r, and Zitnick]{coco}
Tsung-Yi Lin, Michael Maire, Serge Belongie, James Hays, Pietro Perona, Deva Ramanan, Piotr Doll{\'a}r, and C~Lawrence Zitnick.
\newblock Microsoft coco: Common objects in context.
\newblock In \emph{ECCV}, pp.\  740--755, 2014.

\bibitem[Liu et~al.(2024)Liu, Li, Wu, and Lee]{llava}
Haotian Liu, Chunyuan Li, Qingyang Wu, and Yong~Jae Lee.
\newblock Visual instruction tuning.
\newblock \emph{Advances in neural information processing systems}, 36, 2024.

\bibitem[OpenAI(2023)]{openai2023gpt4}
OpenAI.
\newblock {GPT}-4 technical report.
\newblock \emph{arXiv preprint arXiv:2303.08774}, 2023.

\bibitem[Pan et~al.(2023)Pan, Sun, Ge, Li, Duan, Wu, Zhang, Zhou, Qin, Wang, et~al.]{journeydb}
Junting Pan, Keqiang Sun, Yuying Ge, Hao Li, Haodong Duan, Xiaoshi Wu, Renrui Zhang, Aojun Zhou, Zipeng Qin, Yi~Wang, et~al.
\newblock Journeydb: A benchmark for generative image understanding.
\newblock \emph{arXiv preprint arXiv:2307.00716}, 2023.

\bibitem[Peebles \& Xie(2023)Peebles and Xie]{dit}
William Peebles and Saining Xie.
\newblock Scalable diffusion models with transformers.
\newblock In \emph{CVPR}, pp.\  4195--4205, 2023.

\bibitem[Qu et~al.(2024)Qu, Zhang, Liu, Wang, Jiang, Gao, Ye, Du, Yuan, and Wu]{tokenflow}
Liao Qu, Huichao Zhang, Yiheng Liu, Xu~Wang, Yi~Jiang, Yiming Gao, Hu~Ye, Daniel~K Du, Zehuan Yuan, and Xinglong Wu.
\newblock Tokenflow: Unified image tokenizer for multimodal understanding and generation.
\newblock \emph{arXiv preprint arXiv:2412.03069}, 2024.

\bibitem[Radford et~al.(2019)Radford, Wu, Child, Luan, Amodei, Sutskever, et~al.]{gpt2}
Alec Radford, Jeffrey Wu, Rewon Child, David Luan, Dario Amodei, Ilya Sutskever, et~al.
\newblock Language models are unsupervised multitask learners.
\newblock \emph{OpenAI Blog}, 2019.

\bibitem[Rombach et~al.(2022{\natexlab{a}})Rombach, Blattmann, Lorenz, Esser, and Ommer]{ldm}
Robin Rombach, Andreas Blattmann, Dominik Lorenz, Patrick Esser, and Bj{\"o}rn Ommer.
\newblock High-resolution image synthesis with latent diffusion models.
\newblock In \emph{CVPR}, pp.\  10684--10695, 2022{\natexlab{a}}.

\bibitem[Rombach et~al.(2022{\natexlab{b}})Rombach, Blattmann, Lorenz, Esser, and Ommer]{stablediffusion}
Robin Rombach, Andreas Blattmann, Dominik Lorenz, Patrick Esser, and Bj{\"{o}}rn Ommer.
\newblock High-resolution image synthesis with latent diffusion models.
\newblock In \emph{CVPR}, pp.\  10674--10685, 2022{\natexlab{b}}.

\bibitem[Salimans et~al.(2016)Salimans, Goodfellow, Zaremba, Cheung, Radford, and Chen]{is}
Tim Salimans, Ian Goodfellow, Wojciech Zaremba, Vicki Cheung, Alec Radford, and Xi~Chen.
\newblock Improved techniques for training gans.
\newblock In \emph{NeurIPS}, volume~29, 2016.

\bibitem[Schuhmann \& Beaumont(2022)Schuhmann and Beaumont]{laion_aesthetics}
Christoph Schuhmann and Romain Beaumont.
\newblock Laion-aesthetics.
\newblock \url{https://laion.ai/blog/laion-aesthetics/}, 2022.

\bibitem[Sharma et~al.(2018)Sharma, Ding, Goodman, and Soricut]{cc3m}
Piyush Sharma, Nan Ding, Sebastian Goodman, and Radu Soricut.
\newblock Conceptual captions: A cleaned, hypernymed, image alt-text dataset for automatic image captioning.
\newblock In \emph{ACL}, pp.\  2556--2565, 2018.

\bibitem[Shazeer(2020)]{glu}
Noam Shazeer.
\newblock Glu variants improve transformer.
\newblock \emph{arXiv preprint arXiv:2002.05202}, 2020.

\bibitem[Soomro(2012)]{ucf101}
K~Soomro.
\newblock Ucf101: A dataset of 101 human actions classes from videos in the wild.
\newblock \emph{arXiv preprint arXiv:1212.0402}, 2012.

\bibitem[Su et~al.(2024)Su, Ahmed, Lu, Pan, Bo, and Liu]{rope}
Jianlin Su, Murtadha Ahmed, Yu~Lu, Shengfeng Pan, Wen Bo, and Yunfeng Liu.
\newblock Roformer: Enhanced transformer with rotary position embedding, 2024.

\bibitem[Sun et~al.(2024)Sun, Jiang, Chen, Zhang, Peng, Luo, and Yuan]{llamagen}
Peize Sun, Yi~Jiang, Shoufa Chen, Shilong Zhang, Bingyue Peng, Ping Luo, and Zehuan Yuan.
\newblock Autoregressive model beats diffusion: Llama for scalable image generation.
\newblock \emph{arXiv preprint arXiv:2406.06525}, 2024.

\bibitem[Tan et~al.(2024)Tan, Xue, Jia, Wang, Ye, Shi, Sun, Wu, Chen, and Jiang]{sweettokenizer}
Zhentao Tan, Ben Xue, Jian Jia, Junhao Wang, Wencai Ye, Shaoyun Shi, Mingjie Sun, Wenjin Wu, Quan Chen, and Peng Jiang.
\newblock Sweettokenizer: Semantic-aware spatial-temporal tokenizer for compact visual discretization.
\newblock \emph{arXiv preprint arXiv:2412.10443}, 2024.

\bibitem[Team(2024)]{chameleon}
Chameleon Team.
\newblock Chameleon: Mixed-modal early-fusion foundation models.
\newblock \emph{arXiv preprint arXiv:2405.09818}, 2024.

\bibitem[Tian et~al.(2024)Tian, Jiang, Yuan, Peng, and Wang]{var}
Keyu Tian, Yi~Jiang, Zehuan Yuan, Bingyue Peng, and Liwei Wang.
\newblock Visual autoregressive modeling: Scalable image generation via next-scale prediction.
\newblock \emph{arXiv preprint arXiv:2404.02905}, 2024.

\bibitem[Touvron et~al.(2023)Touvron, Martin, Stone, Albert, Almahairi, Babaei, Bashlykov, Batra, Bhargava, Bhosale, Bikel, Blecher, Canton{-}Ferrer, Chen, Cucurull, Esiobu, Fernandes, Fu, Fu, Fuller, Gao, Goswami, Goyal, Hartshorn, Hosseini, Hou, Inan, Kardas, Kerkez, Khabsa, Kloumann, Korenev, Koura, Lachaux, Lavril, Lee, Liskovich, Lu, Mao, Martinet, Mihaylov, Mishra, Molybog, Nie, Poulton, Reizenstein, Rungta, Saladi, Schelten, Silva, Smith, Subramanian, Tan, Tang, Taylor, Williams, Kuan, Xu, Yan, Zarov, Zhang, Fan, Kambadur, Narang, Rodriguez, Stojnic, Edunov, and Scialom]{llama2}
Hugo Touvron, Louis Martin, Kevin Stone, Peter Albert, Amjad Almahairi, Yasmine Babaei, Nikolay Bashlykov, Soumya Batra, Prajjwal Bhargava, Shruti Bhosale, Dan Bikel, Lukas Blecher, Cristian Canton{-}Ferrer, Moya Chen, Guillem Cucurull, David Esiobu, Jude Fernandes, Jeremy Fu, Wenyin Fu, Brian Fuller, Cynthia Gao, Vedanuj Goswami, Naman Goyal, Anthony Hartshorn, Saghar Hosseini, Rui Hou, Hakan Inan, Marcin Kardas, Viktor Kerkez, Madian Khabsa, Isabel Kloumann, Artem Korenev, Punit~Singh Koura, Marie{-}Anne Lachaux, Thibaut Lavril, Jenya Lee, Diana Liskovich, Yinghai Lu, Yuning Mao, Xavier Martinet, Todor Mihaylov, Pushkar Mishra, Igor Molybog, Yixin Nie, Andrew Poulton, Jeremy Reizenstein, Rashi Rungta, Kalyan Saladi, Alan Schelten, Ruan Silva, Eric~Michael Smith, Ranjan Subramanian, Xiaoqing~Ellen Tan, Binh Tang, Ross Taylor, Adina Williams, Jian~Xiang Kuan, Puxin Xu, Zheng Yan, Iliyan Zarov, Yuchen Zhang, Angela Fan, Melanie Kambadur, Sharan Narang, Aur{\'{e}}lien Rodriguez, Robert Stojnic, Sergey Edunov,
  and Thomas Scialom.
\newblock Llama 2: Open foundation and fine-tuned chat models.
\newblock \emph{arXiv preprint arXiv:2307.09288}, 2023.

\bibitem[Tseng et~al.(2021)Tseng, Jiang, Liu, Yang, and Yang]{lecam}
Hung{-}Yu Tseng, Lu~Jiang, Ce~Liu, Ming{-}Hsuan Yang, and Weilong Yang.
\newblock Regularizing generative adversarial networks under limited data.
\newblock In \emph{CVPR}, pp.\  7921--7931, 2021.

\bibitem[Unterthiner et~al.(2018)Unterthiner, Van~Steenkiste, Kurach, Marinier, Michalski, and Gelly]{fvd}
Thomas Unterthiner, Sjoerd Van~Steenkiste, Karol Kurach, Raphael Marinier, Marcin Michalski, and Sylvain Gelly.
\newblock Towards accurate generative models of video: A new metric \& challenges.
\newblock \emph{arXiv preprint arXiv:1812.01717}, 2018.

\bibitem[Van Den~Oord et~al.(2017)Van Den~Oord, Vinyals, et~al.]{vqvae}
Aaron Van Den~Oord, Oriol Vinyals, et~al.
\newblock Neural discrete representation learning.
\newblock volume~30, 2017.

\bibitem[Vaswani et~al.(2017)Vaswani, Shazeer, Parmar, Uszkoreit, Jones, Gomez, Kaiser, and Polosukhin]{attention}
Ashish Vaswani, Noam Shazeer, Niki Parmar, Jakob Uszkoreit, Llion Jones, Aidan~N. Gomez, Lukasz Kaiser, and Illia Polosukhin.
\newblock Attention is all you need.
\newblock In Isabelle Guyon, Ulrike von Luxburg, Samy Bengio, Hanna~M. Wallach, Rob Fergus, S.~V.~N. Vishwanathan, and Roman Garnett (eds.), \emph{NeurIPS}, pp.\  5998--6008, 2017.

\bibitem[Wang et~al.(2024{\natexlab{a}})Wang, Suri, Ren, Chen, and Shrivastava]{larp}
Hanyu Wang, Saksham Suri, Yixuan Ren, Hao Chen, and Abhinav Shrivastava.
\newblock Larp: Tokenizing videos with a learned autoregressive generative prior.
\newblock \emph{arXiv preprint arXiv:2410.21264}, 2024{\natexlab{a}}.

\bibitem[Wang et~al.(2024{\natexlab{b}})Wang, Jiang, Yuan, Peng, Wu, and Jiang]{omnitokenizer}
Junke Wang, Yi~Jiang, Zehuan Yuan, Binyue Peng, Zuxuan Wu, and Yu-Gang Jiang.
\newblock Omnitokenizer: A joint image-video tokenizer for visual generation.
\newblock \emph{arXiv preprint arXiv:2406.09399}, 2024{\natexlab{b}}.

\bibitem[Xie et~al.(2024)Xie, Mao, Bai, Zhang, Wang, Lin, Gu, Chen, Yang, and Shou]{showo}
Jinheng Xie, Weijia Mao, Zechen Bai, David~Junhao Zhang, Weihao Wang, Kevin~Qinghong Lin, Yuchao Gu, Zhijie Chen, Zhenheng Yang, and Mike~Zheng Shou.
\newblock Show-o: One single transformer to unify multimodal understanding and generation.
\newblock \emph{arXiv preprint arXiv:2408.12528}, 2024.

\bibitem[Yu et~al.(2022)Yu, Li, Koh, Zhang, Pang, Qin, Ku, Xu, Baldridge, and Wu]{vit-vqgan}
Jiahui Yu, Xin Li, Jing~Yu Koh, Han Zhang, Ruoming Pang, James Qin, Alexander Ku, Yuanzhong Xu, Jason Baldridge, and Yonghui Wu.
\newblock Vector-quantized image modeling with improved {VQGAN}.
\newblock In \emph{ICLR}, 2022.

\bibitem[Yu et~al.(2023)Yu, Cheng, Sohn, Lezama, Zhang, Chang, Hauptmann, Yang, Hao, Essa, and Jiang]{magvit1}
Lijun Yu, Yong Cheng, Kihyuk Sohn, Jos{\'{e}} Lezama, Han Zhang, Huiwen Chang, Alexander~G. Hauptmann, Ming{-}Hsuan Yang, Yuan Hao, Irfan Essa, and Lu~Jiang.
\newblock {MAGVIT:} masked generative video transformer.
\newblock In \emph{CVPR}, pp.\  10459--10469, 2023.

\bibitem[Yu et~al.(2024{\natexlab{a}})Yu, Lezama, Gundavarapu, Versari, Sohn, Minnen, Cheng, Gupta, Gu, Hauptmann, Gong, Yang, Essa, Ross, and Jiang]{magvit2}
Lijun Yu, Jose Lezama, Nitesh~Bharadwaj Gundavarapu, Luca Versari, Kihyuk Sohn, David Minnen, Yong Cheng, Agrim Gupta, Xiuye Gu, Alexander~G Hauptmann, Boqing Gong, Ming-Hsuan Yang, Irfan Essa, David~A Ross, and Lu~Jiang.
\newblock Language model beats diffusion - tokenizer is key to visual generation.
\newblock In \emph{ICLR}, 2024{\natexlab{a}}.

\bibitem[Yu et~al.(2024{\natexlab{b}})Yu, Weber, Deng, Shen, Cremers, and Chen]{titok}
Qihang Yu, Mark Weber, Xueqing Deng, Xiaohui Shen, Daniel Cremers, and Liang-Chieh Chen.
\newblock An image is worth 32 tokens for reconstruction and generation.
\newblock \emph{arXiv preprint arXiv:2406.07550}, 2024{\natexlab{b}}.

\bibitem[Zhang et~al.(2018)Zhang, Isola, Efros, Shechtman, and Wang]{lpips}
Richard Zhang, Phillip Isola, Alexei~A. Efros, Eli Shechtman, and Oliver Wang.
\newblock The unreasonable effectiveness of deep features as a perceptual metric.
\newblock In \emph{CVPR}, pp.\  586--595, 2018.

\bibitem[Zhang et~al.(2022)Zhang, Roller, Goyal, Artetxe, Chen, Chen, Dewan, Diab, Li, Lin, Mihaylov, Ott, Shleifer, Shuster, Simig, Koura, Sridhar, Wang, and Zettlemoyer]{rmsnorm}
Susan Zhang, Stephen Roller, Naman Goyal, Mikel Artetxe, Moya Chen, Shuohui Chen, Christopher Dewan, Mona~T. Diab, Xian Li, Xi~Victoria Lin, Todor Mihaylov, Myle Ott, Sam Shleifer, Kurt Shuster, Daniel Simig, Punit~Singh Koura, Anjali Sridhar, Tianlu Wang, and Luke Zettlemoyer.
\newblock {OPT:} open pre-trained transformer language models.
\newblock \emph{arXiv preprint arXiv:2205.01068}, 2022.

\end{thebibliography}
\bibliographystyle{iclr2024_conference}


\end{document}